\documentclass[sigconf,nonacm]{acmart}
\AtBeginDocument{%
  }

\newcommand{\cmark}{\ensuremath{\checkmark}}
\newcommand{\xmark}{\ensuremath{\times}}
\begin{document}

\title{IterCAD: Iterative Program Repair for CAD Code Generation from Orthographic Views}

\author{Yuchuan Wu}
\email{ycwu24@m.fudan.edu.cn}
\affiliation{%
  \institution{Fudan University}
  \city{Shanghai}
  \country{China}
}

\author{Ke Niu}
\email{kniu22@m.fudan.edu.cn}
\affiliation{%
  \institution{Fudan University}
  \city{Shanghai}
  \country{China}
}

\author{Haiyang Yu}
\email{hyyu20@fudan.edu.cn}
\affiliation{%
  \institution{Fudan University}
  \city{Shanghai}
  \country{China}
}

\author{Zhuofan Chen}
\email{zfchen23@m.fudan.edu.cn}
\affiliation{%
  \institution{Fudan University}
  \city{Shanghai}
  \country{China}
}

\author{Xiangyang Xue}
\email{xyxue@fudan.edu.cn}
\affiliation{%
  \institution{Fudan University}
  \city{Shanghai}
  \country{China}
}

\author{Bin Li}
\correspondingauthor
\email{libin@fudan.edu.cn}
\affiliation{%
  \institution{Fudan University}
  \city{Shanghai}
  \country{China}
}

\renewcommand{\shortauthors}{Yuchuan Wu et al.}


\begin{abstract}
Generating executable CAD code from dimension-annotated orthographic drawings is a challenging task requiring geometric understanding, procedural reasoning, and precise numerical prediction. Existing vision-language approaches typically formulate this problem as one-shot generation, preventing the model from inspecting intermediate CAD results and correcting early mistakes, often leading to non-executable code or geometrically inconsistent outputs. In this paper, we propose \textbf{IterCAD}, an iterative framework that reformulates orthographic-view-to-CAD generation as a progressive program repair process. Instead of predicting the final CAD code in a single pass, IterCAD repeatedly analyzes the current CAD result, reasons about its discrepancy with the target views, and explicitly decides whether to \texttt{REVISE} the code or \texttt{STOP} the refinement process. To make iterative repair learnable, we further construct \textbf{IterCAD-RS}, a structured revise-or-stop supervision set containing both repairable intermediate CAD states and already-correct states, and develop a three-stage training strategy for initial generation, revision learning, and multi-turn RL optimization. By closing the loop between visual understanding, geometric verification, and code refinement, IterCAD progressively corrects structural and parametric errors. Experiments on CADExpert show that IterCAD consistently improves code executability and geometric fidelity over strong one-shot baselines.
\end{abstract}


\maketitle

\begin{figure*}[t]
  \centering
  \includegraphics[width=\textwidth]{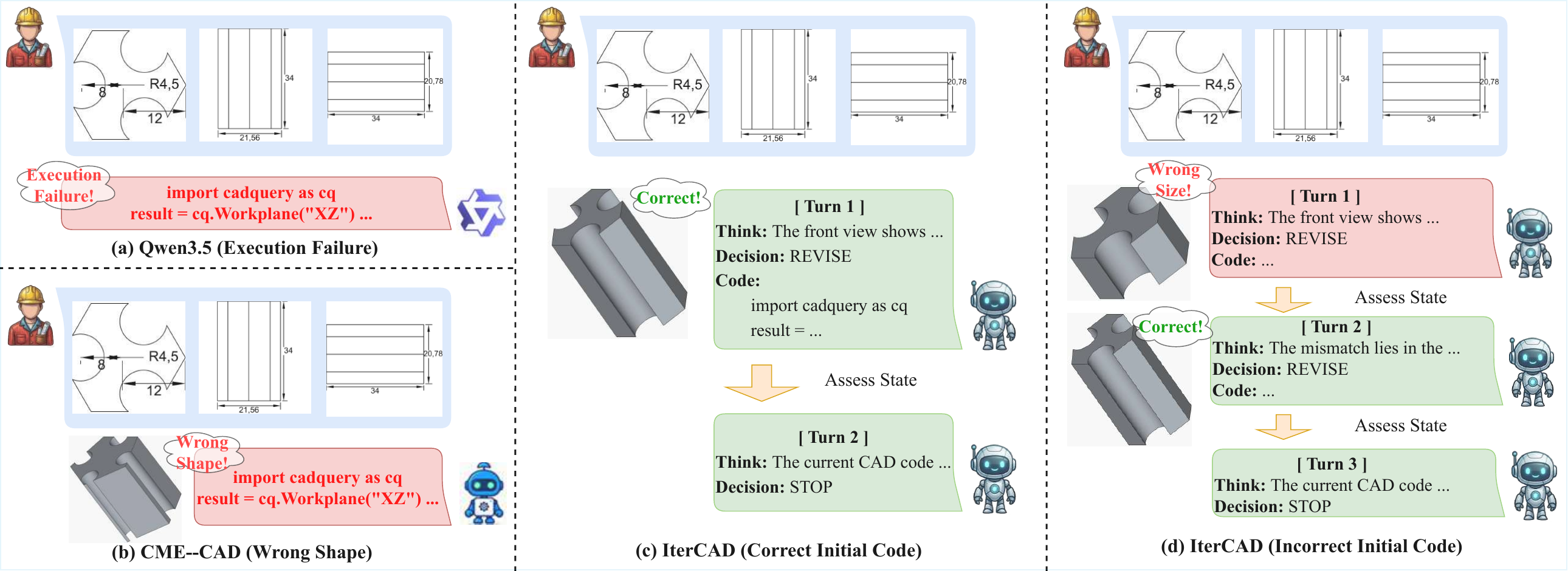}
  \caption{Comparison of one-shot CAD generation and \textbf{IterCAD}.
  While one-shot methods often fail to produce correct CAD programs,
  IterCAD progressively improves the program through iterative state
  assessment and revision.}
  \Description{Comparison between one-shot CAD generation methods and
  IterCAD. One-shot methods may produce execution failures or incorrect
  geometry, while IterCAD repeatedly assesses and revises the current CAD
  program until obtaining a correct result.}
  \label{fig:teaser}
\end{figure*}

\section{Introduction}

As a fundamental tool in industrial design and manufacturing, Computer-Aided Design (CAD)~\cite{sun2025large} has long been essential for the digital creation of engineering products through formalized modeling languages. In modern production pipelines, design concepts are typically translated into precise and editable CAD models before fabrication, simulation, and downstream verification. As a result, automatically generating high-quality CAD code has become an important research problem with both academic significance and practical value.

Compared with directly reconstructing low-level geometric representations such as point clouds, voxels, or meshes, generating parametric CAD code~\cite{wu2021deepcad, seff2021vitruvion} is more aligned with real-world engineering requirements. CAD code is compact, editable, and semantically meaningful: it explicitly encodes the construction logic of a shape, supports downstream modification and validation, and can be directly integrated into industrial design workflows. Therefore, CAD code generation is not merely a 3D reconstruction problem, but a structured code generation task that requires both geometric understanding and procedural reasoning.

From the perspective of practical design workflows, generating CAD code from orthographic engineering drawings is particularly meaningful~\cite{wang20252d}. In real industrial scenarios, engineers commonly communicate geometric intent through orthographic projections with precise dimension annotations, rather than through free-form natural language descriptions. Such drawings naturally provide the structural and numerical information required for CAD modeling, making them a more realistic and scalable input modality for automatic CAD generation. However, converting dimension-annotated orthographic views into executable CAD code remains highly challenging, since the model must simultaneously infer the underlying 3D geometry, recover the correct sequence of modeling operations, and predict precise numerical parameters.

Existing vision-language approaches~\cite{wang20252d,niu2025cme,niu2026intent} have shown promising progress by directly mapping input drawings to CAD code in a one-shot manner. However, this paradigm suffers from an inherent limitation for orthographic-view-to-CAD generation. Generating correct CAD code is a strongly coupled sequential process: an early mistake in workplane selection, operation ordering, geometric topology, or dimensional prediction can propagate through subsequent steps and lead to invalid or geometrically inconsistent results. More importantly, once such an error is made, one-shot methods have no opportunity to inspect the generated result, identify the discrepancy with the target drawing, and revise the code accordingly. As shown in Figure~\ref{fig:teaser}, existing one-shot methods may produce either non-executable code or incorrect geometry, while many failures in fact remain repairable if the model is allowed to check the current CAD result and refine it iteratively.

We argue that this limitation is especially severe in the setting of dimension-annotated orthographic-view-to-CAD generation. Unlike tasks where approximate outputs may still be acceptable, CAD code generation demands strict executability and geometric faithfulness. A small deviation in a hole diameter, extrusion depth, or Boolean operation may render the code unusable or significantly alter the final shape. At the same time, CAD code generation also presents a favorable property for iterative refinement: intermediate outputs take the form of executable CAD code that can be inspected, compared against the target views, and further revised. This suggests that CAD code generation should not be treated as a single-pass prediction problem, but rather as a progressive refinement process in which the model can repeatedly examine its current result and correct its own mistakes.

Based on this insight, we propose \textbf{IterCAD}, an iterative framework for CAD code generation from dimension-annotated orthographic drawings. Instead of producing the final answer in one step, \textbf{IterCAD} enables the model to repeatedly analyze the current CAD result, reason about its mismatch with the target views, and explicitly decide whether to \texttt{REVISE} the code or \texttt{STOP} the refinement process. To make such iterative repair learnable, we further construct \textbf{IterCAD-RS}, a structured revise-or-stop supervision set, and develop a dedicated three-stage training recipe for initial generation, revision learning, and multi-turn policy optimization. By closing the loop between visual understanding, geometric verification, and code revision, \textbf{IterCAD} progressively corrects both structural and parametric errors, leading to more accurate and executable CAD reconstruction.

Different from simply eliciting longer reasoning traces, \textbf{IterCAD} introduces a structured iterative paradigm tailored to the characteristics of CAD modeling. It allows the model to recover from erroneous initial generations, refine nearly correct CAD code with subtle dimensional inconsistencies, and terminate the process explicitly once sufficient consistency has been achieved. Extensive experiments demonstrate that \textbf{IterCAD} consistently improves both executability and geometric accuracy over strong one-shot baselines.

\begin{table*}[t]
\centering
\caption{Perturbation taxonomy used to construct \textbf{IterCAD-RS}. Starting from a correct structured CAD state (\emph{ShapeSpec}), we inject one or two geometric perturbations to synthesize executable but geometrically inconsistent intermediate states for revise-or-stop supervision.}
\label{tab:perturbation_taxonomy}
\resizebox{\textwidth}{!}{
\begin{tabular}{llll}
\toprule
\textbf{Structural Family} & \textbf{Perturbation Target} & \textbf{Representative Operations} & \textbf{Typical Resulting Errors} \\
\midrule
Body-level
& Sketch plane
& Change \texttt{XY}/\texttt{YZ}/\texttt{XZ} workplane
& Wrong global orientation or projection mismatch \\

Body-level
& Base profile geometry
& Modify primitive type, polygon side count, or profile scale
& Incorrect outer silhouette or overall geometry \\

Body-level
& Extrusion parameter
& Modify extrusion height or depth
& Incorrect thickness or global dimensions \\

\midrule
\texttt{cut}
& Inner profile geometry
& Change inner profile type or size
& Wrong cavity shape or incorrect inner dimensions \\

\texttt{cut}
& Modifier existence
& Remove the cut operation
& Missing inner structure \\

\midrule
\texttt{hole\_pattern}
& Hole multiplicity
& Increase or decrease the number of holes
& Incorrect hole count \\

\texttt{hole\_pattern}
& Hole layout
& Modify radial offset or distribution radius
& Wrong hole arrangement \\

\texttt{hole\_pattern}
& Hole size / existence
& Change hole diameter or remove the pattern
& Incorrect or missing hole structure \\

\midrule
\texttt{edge\_finish}
& Finish type
& Switch between \texttt{fillet} and \texttt{chamfer}
& Incorrect edge treatment \\

\texttt{edge\_finish}
& Finish magnitude / existence
& Modify finish value or remove the operation
& Wrong or missing local edge detail \\

\midrule
Topology-level
& Modifier insertion
& Add an extra \texttt{cut}, \texttt{hole\_pattern}, or \texttt{edge\_finish}
& Spurious geometric structure \\
\bottomrule
\end{tabular}
}
\end{table*}

Our main contributions are summarized as follows:
\begin{itemize}
    \item We reformulate dimension-annotated orthographic-view-to-CAD generation as an iterative program repair problem, in which the model operates on executable intermediate CAD states and learns to make explicit \texttt{REVISE}/\texttt{STOP} decisions. To support this formulation, we construct \textbf{IterCAD-RS}, a structured revise-or-stop supervision set containing both repairable CAD states paired with \texttt{REVISE} targets and already-correct CAD states paired with \texttt{STOP} targets.
    \item We propose \textbf{IterCAD}, a unified iterative framework that integrates discrepancy-aware reasoning, explicit revise-or-stop decision-making, and iterative CAD code refinement in a closed loop. Building on this framework, we develop a dedicated three-stage training recipe that progressively teaches the model initial CAD generation, intermediate-state revision, and multi-turn self-repair policy optimization.
    \item Extensive experiments on CADExpert~\cite{niu2025cme} show that \textbf{IterCAD} consistently outperforms strong one-shot baselines in both code executability and geometric fidelity. Further analyses verify the effectiveness of the iterative program repair formulation, \textbf{IterCAD-RS}, and the proposed three-stage training strategy.
\end{itemize}

\section{Related Work}

\subsection{CAD Code Generation from Visual Inputs}

Large vision-language models have demonstrated substantial value
across diverse application domains~\cite{peng2025interpretable,
fu2026omnipt,Niu_2025_ICCV,fu2023denoising,yu2025umitunifyingmedicalimaging,yu2025eveendtoendvideosubtitle,10376764}. Building on these advances, recent
research has increasingly explored their potential for CAD code
generation. CAD code generation aims to produce executable and
editable 3D design programs directly from user inputs, such as text
descriptions, images, or engineering drawings~\cite{xu2024cad,
alrashedy2024generating,guan2025cad,alam2024gencad,
qin2025drawing2cad}. Compared with general code generation, this task
requires not only structured program prediction but also accurate
spatial reasoning, geometric consistency, and numerical precision.
Even small structural or parametric errors may lead to non-executable
programs or geometries that deviate substantially from the target
design.

Recent studies have explored CAD generation from visual inputs under different formulations. CAD-MLLM~\cite{xu2024cad} and GenCAD~\cite{alam2024gencad} use vision-language models to directly translate visual inputs into CAD commands or programs, enabling end-to-end visual-to-code generation. Img2CAD~\cite{chen2025img2cad} adopts a two-stage framework that decouples structure prediction from parameter regression, improving flexibility in program synthesis. CAD2Program~\cite{wang20252d} introduces a more expressive code representation to better support complex CAD construction. CAD-Llama~\cite{li2025cad} and CAD-Coder~\cite{guan2025cad} further improve structured parametric CAD generation through hierarchical annotations or expert-designed intermediate descriptions.

Beyond direct visual-to-code prediction, some works have begun to incorporate verification or improvement mechanisms into CAD generation. In particular, CADCodeVerify~\cite{alrashedy2024generating} introduces an automated verification-and-improvement pipeline in which a pre-trained vision-language model inspects the rendered CAD result, generates and answers validation questions, and feeds the resulting feedback back to the generator for refinement. This line of work shows that CAD generation can benefit from going beyond pure one-shot prediction, especially since generated CAD programs are structured, executable, and amenable to downstream checking.

Nevertheless, existing approaches still mainly treat refinement as an external feedback step applied to the generated final program, rather than explicitly modeling CAD generation itself as a learned multi-turn repair process over intermediate executable states. In contrast to CADCodeVerify, which relies on prompting-based automated verification at inference time, our method adopts a unified iterative formulation in which the model directly operates on intermediate CAD states, progressively revises them, and explicitly learns a revise-or-stop policy for adaptive refinement.

\subsection{Reinforcement Learning in CAD Generation}

Reinforcement learning~\cite{sutton1988learning,watkins1992q,sutton1999policy,schulman2015trust,schulman2017proximal,mnih2016asynchronous,jia2025memlgrpoheterogeneousmultiexpertmutual} has recently emerged as a promising direction for improving CAD code generation beyond standard supervised fine-tuning~\cite{niu2025creft}. Instead of relying solely on next-token prediction, RL-based methods optimize generated CAD programs with task-specific objectives, such as executability, geometric accuracy, and alignment with design intent.

Among existing works, CAD-RL~\cite{niu2026intent} is a representative effort in this direction. It introduces a multimodal chain-of-thought-guided reinforcement learning framework for precise CAD code generation, showing that reward-driven optimization can substantially improve the numerical accuracy, executability, and reasoning quality of generated CadQuery programs. This line of work highlights the value of incorporating execution-aware and geometry-aware feedback into CAD generation.

However, existing RL-based CAD generation methods still primarily optimize the quality of the final generated program, rather than explicitly modeling generation as a multi-turn repair process over intermediate executable CAD states. In contrast, our method focuses on iterative program revision, where the model repeatedly inspects the current CAD result, decides whether to \texttt{REVISE} or \texttt{STOP}, and performs adaptive multi-step correction accordingly.

\section{Iterative Repair Formulation and IterCAD-RS Construction}

Orthographic-view-to-CAD generation is naturally suited to an iterative program repair paradigm, since intermediate CAD code is executable and can be further revised based on discrepancies with the target orthographic views. In this section, we first formulate the task as a multi-turn revise-or-stop process, and then describe the construction of \textbf{IterCAD-RS}, a structured revise-or-stop supervision set for iterative program repair.

\subsection{Multi-turn Program Repair Formulation}

We formulate orthographic-view-to-CAD generation as an iterative program repair process rather than a one-shot prediction problem. Given the input orthographic views $x$, the model progressively refines CAD code over multiple turns until it decides to stop. The initial state is defined as
\begin{equation}
s_0 = (x, \texttt{EMPTY}),
\end{equation}
where no CAD code has been generated yet. At turn $t$, the state is defined as
\begin{equation}
s_t = (x, h_{t-1}),
\end{equation}
where $h_{t-1}$ denotes the interaction history available before turn $t$. In our setting, this history contains the model outputs from previous turns, including the generated \texttt{<think>} rationale, the \texttt{<decision>}, and the current cad \texttt{<code>}. Therefore, the model does not merely revise the latest CAD code in isolation, but performs the next repair step by conditioning on its own earlier reasoning process, decisions, and current code state.

At each turn $t$, the model outputs a structured response
\begin{equation}
\label{ot}
o_t = (r_t, a_t, y_t),
\end{equation}
where $r_t$ denotes the \texttt{<think>} rationale, $a_t \in \{\texttt{REVISE}, \texttt{STOP}\}$ is the decision in the \texttt{<decision>} field, and $y_t$ is the CAD code in the \texttt{<code>} field. The interaction history is updated as
\begin{equation}
h_t = h_{t-1} \oplus o_t,
\end{equation}
where $\oplus$ denotes appending the current-turn output to the history. If $a_t=\texttt{REVISE}$, the updated CAD code $y_t$ is used as the input state for the next turn; if $a_t=\texttt{STOP}$, the iterative process terminates and $y_t$ is taken as the final prediction. The trajectory terminates when
\begin{equation}
a_t = \texttt{STOP}
\quad \text{or} \quad
t = T_{\max}.
\end{equation}

This formulation highlights several important properties of our setting. First, intermediate outputs are structured and executable intermediate states rather than transient text tokens, and can therefore support subsequent refinement. Second, stopping is modeled as a learned action instead of a hand-crafted post-processing rule, enabling the model to adapt the number of refinement steps to the difficulty of each sample. More broadly, this formulation internalizes the core steps of CAD generation---reasoning over the input views, evaluating the current code state, deciding whether further correction is needed, and producing the next revision---within a single unified model, rather than relying on external verification tools or manually designed control heuristics. In this sense, \textsc{IterCAD} goes beyond one-shot CAD code prediction toward a more adaptive and self-refining paradigm for CAD code generation.

\subsection{IterCAD-RS: Structured Revise-or-Stop Supervision for Iterative Program Repair}

A key supervision bottleneck for iterative program repair is that standard orthographic-view-to-CAD datasets provide only the final correct CAD code, but not the intermediate states needed for learning revise-or-stop behavior. To address this, we construct \textbf{IterCAD-RS}, a structured supervision set built on top of the SFT training split of CADExpert~\cite{niu2025cme}, which augments each ground-truth sample with both repairable intermediate states and positive stopping states.

Specifically, instead of perturbing raw code strings directly, we first convert each correct CadQuery program into a structured geometric state representation, denoted as \emph{ShapeSpec}, and then apply controlled semantic perturbations in this space to synthesize executable but geometrically inconsistent intermediate states. These perturbations cover both body-level and modifier-level attributes, producing realistic errors that remain executable while deviating from the target orthographic views, as summarized in Table~\ref{tab:perturbation_taxonomy}. We further apply validity-preserving filtering to remove degenerate or unchanged cases.

For each synthesized wrong state, we pair it with the original correct target and automatically generate a short repair description, denoted as \emph{fix\_text}, which serves as supervision for the \texttt{<think>} field. This yields revision samples of the form
\[
(\text{views}, \text{wrong code}) \rightarrow (\text{repair rationale}, \texttt{REVISE}, \text{correct code}),
\]
while correct or semantically equivalent executable states are used to construct \texttt{STOP} supervision. In this way, \textsc{IterCAD-RS} provides explicit supervision for both intermediate revision and termination behavior. More details on perturbation design, filtering rules, and data statistics are provided in the supplementary material.

\section{Methodology}

\begin{figure*}[t]
    \centering
    \includegraphics[width=\textwidth]{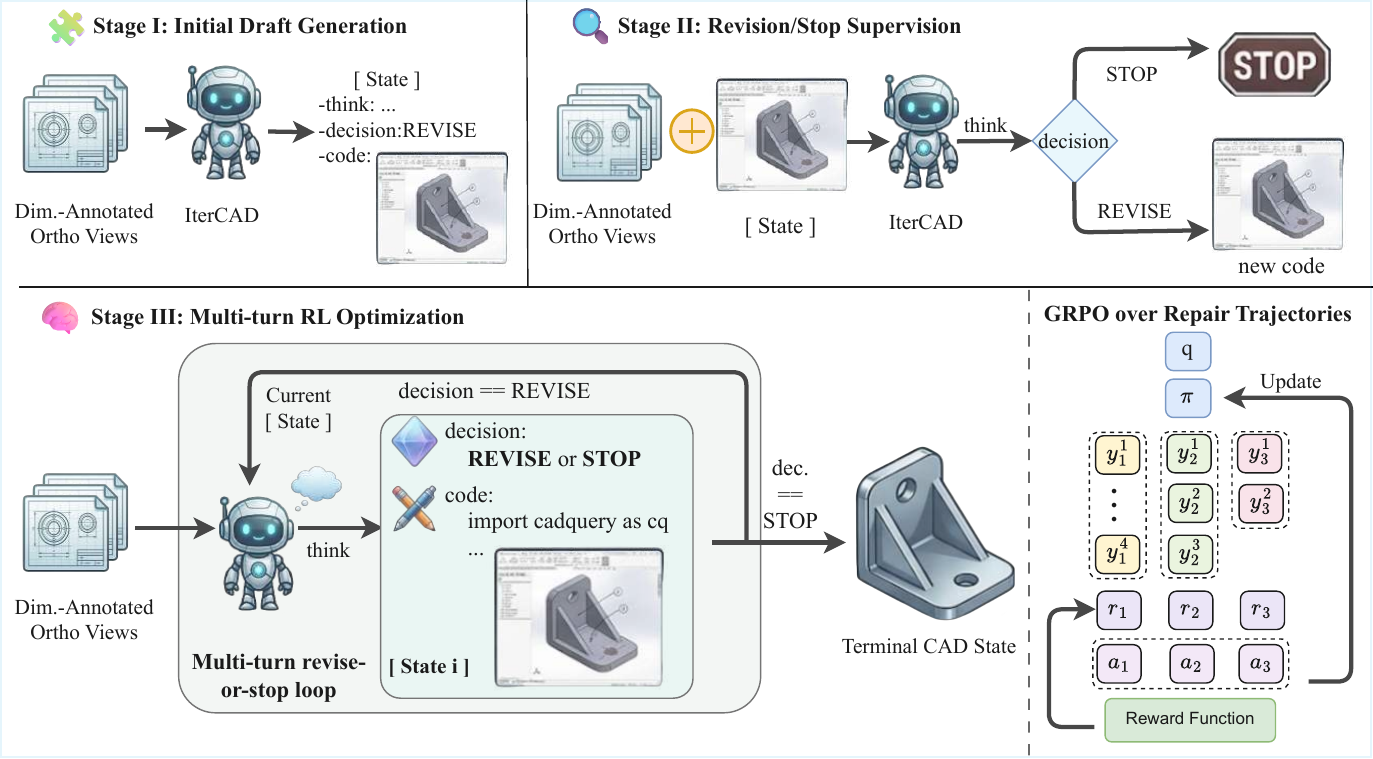}
    \caption{Overview of \textbf{IterCAD}. Stage~I learns first-turn initial draft generation from dimension-annotated orthographic views. Stage~II learns structured revise-or-stop behavior from intermediate CAD states. Stage~III further optimizes the full multi-turn repair process with GRPO over sampled repair trajectories, yielding a unified iterative policy for initial generation, revision, and adaptive stopping.}
    \label{fig:architecture}
\end{figure*}

\subsection{Overview of IterCAD}

As illustrated in Figure~\ref{fig:architecture}, \textbf{IterCAD} formulates orthographic-view-to-CAD generation as an iterative program repair process rather than a one-shot prediction problem. The overall framework consists of three training stages. In Stage~I, the model learns to generate an initial CAD draft from dimension-annotated orthographic views, establishing the basic vision-to-code mapping. In Stage~II, the model learns revise-or-stop behavior from intermediate CAD states: given the target views together with a current CAD state, it reasons about the mismatch between the current CAD code and the target specification, and outputs either a \texttt{STOP} decision if the current code is already satisfactory or a \texttt{REVISE} decision followed by updated CAD code otherwise. In Stage~III, we further optimize the full multi-turn repair process with GRPO over sampled repair trajectories. While the first two stages equip the model with the basic capabilities for initial generation and intermediate revision, they do not by themselves ensure that the model can effectively coordinate these abilities over a multi-turn process. By optimizing over diverse sampled trajectories, Stage~III encourages the model to jointly leverage prior generation and revision knowledge, learn when additional refinement is truly beneficial, and decide when the current state is sufficient to terminate.

\subsection{Training Data}

The three training stages of \textbf{IterCAD} use different but complementary forms of supervision, all built upon CADExpert~\cite{niu2025cme}, an industrial-oriented benchmark for executable and editable CAD code generation. Each sample in CADExpert contains dimension-annotated orthographic views together with the corresponding executable CADQuery code, providing the basic supervision for orthographic-view-to-CAD learning.

In Stage~I, we directly use 8,960 standard orthographic-view-to-CAD pairs from the SFT training split of CADExpert to train the model for initial CAD code generation. In Stage~II, we further construct \textbf{IterCAD-RS} on top of this split, yielding 22,000 structured revise-or-stop supervision samples derived from intermediate CAD states. In Stage~III, we use an additional 4,480 training samples for multi-turn RL optimization. The data used in Stage~III is disjoint from that used in Stages~I and II.

Overall, the training data progresses from direct generation supervision, to structured revise-or-stop supervision, and finally to trajectory-level optimization. More details on data construction are provided in the supplementary material.

\subsection{Three-stage Training Recipe}

As illustrated in Figure~\ref{fig:architecture}, \textbf{IterCAD} is trained with a three-stage recipe rather than end-to-end joint optimization from scratch. This design is motivated by the observation that initial CAD draft generation, revise-or-stop decision making, and full multi-turn self-correction are related but distinct capabilities. While the first two stages provide the model with the basic abilities for initial generation and intermediate repair, they do not by themselves ensure effective coordination of these abilities during iterative inference. Therefore, we progressively train the model from supervised initial drafting, to structured revise-or-stop learning, and finally to trajectory-level RL optimization, yielding a more stable and effective training process for iterative program repair.

\subsubsection{Stage I: Initial Draft Generation}

In Stage~I, we train the model to produce the first-turn structured response from the input orthographic views. Different from standard one-shot CAD code generation, the model does not directly output only the final CAD code. Instead, it is trained under the same interaction format used in later stages, where the current CAD state is initialized as empty and the model performs the first revision step accordingly.

Formally, given the input orthographic views $x$ and an empty initial code state, the model predicts
\begin{equation}
(x, \texttt{EMPTY}) \rightarrow (r_1, \texttt{REVISE}, y_1),
\label{eq:stage1_obj}
\end{equation}
where $r_1$ denotes the \texttt{<think>} rationale and $y_1$ denotes the CAD code in the \texttt{<code>} field. Since no valid CAD code exists at the first turn, the decision is always set to \texttt{REVISE}, and the model is required to generate an initial executable CAD draft.

This stage serves two purposes. First, it establishes the basic vision-to-code capability from dimension-annotated orthographic views. Second, by aligning the output format with the later revise-or-stop stages, it provides a consistent initialization for iterative program repair, so that subsequent stages can directly build on the same structured interaction pattern.

\subsubsection{Stage II: Revision/Stop Supervision}

In Stage~II, we train the model with structured revise-or-stop supervision on intermediate CAD states. Given the target orthographic views $x$ together with the interaction history $h_{t-1}$, the model predicts a structured output
\begin{equation}
(x, h_{t-1}) \rightarrow (r_t, a_t, y_t),
\end{equation}
where $r_t$ denotes the \texttt{<think>} rationale, $a_t \in \{\texttt{REVISE}, \texttt{STOP}\}$ is the decision in the \texttt{<decision>} field, and $y_t$ denotes the CAD code in the \texttt{<code>} field. Here, $h_{t-1}$ contains the model outputs from previous turns, including the generated rationale, decision, and current CAD code. Therefore, the model does not revise the current code in isolation, but predicts the next action by conditioning on its own earlier reasoning process, decisions, and code state.

The supervision used in this stage is provided by \textbf{IterCAD-RS}, which contains two complementary types of samples: revision samples, where executable but geometrically inconsistent intermediate states are paired with repair rationales, \texttt{REVISE} decisions, and corrected CAD code; and stopping samples, where already-correct or semantically equivalent executable states are paired with \texttt{STOP} decisions. In this way, Stage~II teaches the model not only how to perform targeted correction when geometric discrepancies exist, but also how to decide when the current state is already sufficient to terminate.

Importantly, the training targets preserve the same structured output protocol used at inference time, namely \texttt{<think>}, \texttt{<decision>}, and \texttt{<code>}. As a result, Stage~II equips the model with the core abilities required for iterative program repair: interpreting intermediate states, producing explicit revise-or-stop decisions, and generating updated CAD code in a unified format.

\subsubsection{Stage III: Multi-turn RL Optimization}

After the first two supervised stages, the model has acquired the basic abilities required for iterative CAD repair, including initial draft generation, intermediate-state revision, and explicit stopping. However, supervised training alone does not guarantee that the model can effectively coordinate these abilities during multi-turn inference. In particular, while Stage~I and Stage~II teach the model how to generate, revise, and stop under supervised targets, they do not by themselves ensure that these behaviors will be integrated into a coherent repair policy over full trajectories.

To address this, in Stage~III we further optimize the complete multi-turn repair process using GRPO~\cite{shao2024deepseekmath}. Starting from an initial CAD draft, the model interacts with its own intermediate CAD states and produces a repair trajectory
\begin{equation}
\tau = (s_0,o_1,s_1,\ldots,o_T,s_T),
\label{eq:trajectory}
\end{equation}
where each turn output $o_t$ follows Eq.~(\ref{ot}). The Stage~III objective is to maximize the expected trajectory-level reward
\begin{equation}
\max_{\pi_\theta} \ \mathbb{E}_{\tau \sim \pi_\theta}\left[ R(\tau) \right],
\label{eq:stage3_obj}
\end{equation}
where $R(\tau)$ evaluates the overall quality of the multi-turn repair process.

A key property of this stage is that optimization is performed over sampled multi-turn trajectories instead of isolated single-step corrections. This is important in our setting, since \textbf{IterCAD} is intended to learn a revise-or-stop policy over executable intermediate CAD states, rather than a final-step correction heuristic. By optimizing Eq.~(\ref{eq:stage3_obj}) over diverse sampled trajectories, GRPO encourages the model to jointly leverage the generation and revision knowledge acquired in the previous stages, learn when further refinement is beneficial, and decide when the current state is already sufficient to terminate. As a result, Stage~III turns the model from a supervised reviser into a true multi-turn iterative repair policy.

\begin{table}[t]
\centering
\small
\caption{Main results on CADExpert. IterCAD-Q3VL and IterCAD-Q3.5 denote our method instantiated with Qwen3-VL-8B-Instruct and Qwen3.5-9B, respectively. Best results are shown in \textbf{bold}, and second-best results are \underline{underlined}.}
\label{tab:main_results}
\begin{tabular}{lcccc}
\toprule
\textbf{Model} & \textbf{IoU (\%)} $\uparrow$ & \textbf{Mean CD} $\downarrow$ & \textbf{Med CD} $\downarrow$ & \textbf{Exec. (\%)} $\uparrow$ \\
\midrule
LLaVA-1.5~\cite{liu2024improvedbaselinesvisualinstruction}      & 0.73  & 29.36 & 8.14  & 4.79  \\
Phi-3.5-Vision~\cite{abdin2024phi3technicalreporthighly} & 3.44  & 27.92 & 7.75  & 8.50  \\
InternVL2.5~\cite{chen2024expanding}    & 11.98 & 22.27 & 7.36  & 23.92 \\
Qwen2.5-VL~\cite{bai2025qwen25vltechnicalreport}     & 19.21 & 20.68 & 6.64  & 31.27 \\
InternVL3~\cite{zhu2025internvl3exploringadvancedtraining}      & 24.59 & 21.40 & 6.62  & 29.68 \\
Gemini2.5 Pro~\cite{comanici2025gemini} & 30.86 & 14.13 & 5.97  & 39.40 \\
GPT5-Mini~\cite{singh2025openai}      & 35.15 & 9.89  & 4.81  & 47.13 \\
Doubao-1.6~\cite{guo2025seed1}     & 35.62 & 7.54  & 4.35  & 52.89 \\
Qwen3-VL~\cite{bai2025qwen3}       & 37.04 & 6.96  & 3.84  & 54.79 \\
Qwen3.5~\cite{qwen3.5}      & 43.91 & 5.36  & 3.17  & 59.61 \\
CAD-RL~\cite{niu2026intent}   & 71.84 & 1.38  & 0.36  & 97.32 \\
CME-CAD~\cite{niu2025cme}        & 80.71 & 1.00  & 0.11  & \underline{98.25} \\
\midrule
IterCAD-Q3VL   & \underline{85.10} & \underline{0.8564} & \underline{0.1069} & 97.31 \\
IterCAD-Q3.5   & \textbf{91.61} & \textbf{0.5387} & \textbf{0.1038} & \textbf{99.33} \\
\bottomrule
\end{tabular}
\end{table}

\subsection{Reward Design}

To optimize \textbf{IterCAD} under the multi-turn revise-or-stop setting, we design a composite reward that combines final CAD quality, structural validity, and decision correctness. Since the last turn may output \texttt{STOP} without providing a new \texttt{<code>} block, we evaluate the trajectory using an \emph{effective final code}, i.e., the CAD code corresponding to the terminal program state. If the last turn outputs \texttt{REVISE} with a valid \texttt{<code>} block, we use that code directly; otherwise, we trace back to the most recent valid CAD code in the trajectory history.

Based on the effective final code, the overall reward is defined as
\begin{equation}
R =
R_{\text{final}}
+ \lambda_f R_{\text{format}}
+ \lambda_s R_{\text{syntax}}
+ \lambda_d R_{\text{decision}},
\label{eq:reward}
\end{equation}
where $\lambda_f=0.10$, $\lambda_s=0.05$, and $\lambda_d=0.20$. Here, $R_{\text{final}}$ serves as the dominant learning signal, while the remaining terms act as auxiliary regularizers.

Specifically, $R_{\text{format}}$ measures whether the model follows the required structured interaction protocol, encouraging stable outputs in the \texttt{<think>}, \texttt{<decision>}, and \texttt{<code>} format. $R_{\text{syntax}}$ focuses on code validity, rewarding syntactically well-formed executable code while serving only as a lightweight regularizer. $R_{\text{decision}}$ focuses on revise-or-stop behavior, encouraging the model to stop only when the current CAD state is already sufficiently good and discouraging premature stopping or invalid decisions. Finally, $R_{\text{final}}$ evaluates the geometric quality of the effective final code and remains the primary optimization target throughout training.

Overall, this reward is outcome-centric but behavior-aware: it prioritizes the correctness of the terminal CAD state while also regularizing the model to produce valid structured outputs and make appropriate stopping decisions. More detailed definitions and implementation details are provided in the supplementary material.

\section{Experiments}

\subsection{Dataset and Metrics}

We conduct all experiments on CADExpert~\cite{niu2025cme}, an industrial-oriented benchmark for executable and editable CAD code generation, following its standard benchmark setting for orthographic-view-to-CAD generation.

Following prior work~\cite{niu2025cme}, we evaluate model performance using four complementary metrics: Intersection-over-Union (IoU), Mean Chamfer Distance (Mean CD), Median Chamfer Distance (Med CD), and Executability (Exec.). IoU measures volumetric overlap between predicted and reference shapes, Mean CD and Med CD measure geometric discrepancy in point cloud space, and Executability measures whether the generated CAD code can be successfully executed. Higher values are better for IoU and Exec., while lower values are better for Mean CD and Med CD.

\subsection{Implementation Details}

We evaluate \textbf{IterCAD} on two backbone models: \textbf{Qwen3.5-9B}~\cite{qwen3.5} and \textbf{Qwen3-VL-8B-Instruct}~\cite{qwen3technicalreport}. For both backbones, we adopt the same three-stage training recipe, where the Stage~I checkpoint initializes Stage~II, and the Stage~II checkpoint further initializes Stage~III. Stage~I and Stage~II are trained with supervised fine-tuning, while Stage~III is optimized with GRPO. Unless otherwise specified, all three stages use a learning rate of $1\times10^{-5}$ and are trained for one epoch.

For Stage~III, we use $4$ rollouts per training sample, and set the maximum number of refinement turns to $4$. The same rollout setting is used for both backbone models.

At inference time, the model follows the same iterative repair protocol as in training. Given the input orthographic views, it first produces an initial CAD draft and then iteratively predicts either \texttt{REVISE} or \texttt{STOP}. The process terminates when the model outputs \texttt{STOP} or when the maximum number of refinement rounds reaches $4$. No external stopping heuristic is used. For the main benchmark results, both backbone models are evaluated under the same training and inference pipeline.

\subsection{Main Results}

Table~\ref{tab:main_results} reports the main results on CADExpert.
\textbf{IterCAD} achieves the best overall performance, with
IterCAD-Q3.5 ranking first on all four metrics. Specifically, it
reaches 91.61\% IoU, 0.5387 Mean CD, 0.1038 Med CD, and 99.33\%
executability, establishing a new state of the art for
orthographic-view-to-CAD generation.

Compared with the strong multi-expert baseline CME-CAD,
IterCAD-Q3.5 improves IoU from 80.71\% to 91.61\%, reduces Mean CD
from 1.00 to 0.5387, and increases executability from 98.25\% to
99.33\%. While CME-CAD strengthens one-shot generation through
expert decomposition, IterCAD uses a unified model to iteratively
inspect and repair executable intermediate CAD states. The substantial
improvement therefore suggests that explicit self-correction is more
effective than strengthening single-pass prediction alone.

This conclusion is further supported by IterCAD-Q3VL. Despite using
the lighter Qwen3-VL-8B-Instruct backbone, it surpasses CME-CAD on
all geometric metrics, achieving 85.10\% IoU, 0.8564 Mean CD, and
0.1069 Med CD while maintaining high executability. The consistent
gains across two backbones indicate that the improvement primarily
comes from the iterative repair formulation rather than a particular
model architecture. This robustness is particularly important because
the two backbones differ in capacity and pretraining, yet benefit from
the same repair strategy.

Compared with general-purpose pretrained VLMs, both specialized CAD
methods retain a large advantage, confirming the need for
task-specific modeling of geometry, code structure, and executability.
Moreover, the gains span both geometric overlap and execution
reliability, indicating that iterative refinement improves not only
accuracy but also practical usability. Overall, IterCAD improves
geometric fidelity without sacrificing code validity, supporting
multi-turn repair as an effective alternative to one-shot CAD code
generation.

\begin{table}[t]
\centering
\small
\caption{Ablation study of IterCAD on CADExpert. S1, S2, and S3
denote Stages~I, II, and III, respectively, and MT denotes multi-turn
inference. Best results are shown in \textbf{bold}.}
\label{tab:ablation_main}
\setlength{\tabcolsep}{4pt}
\begin{tabular}{cccc|cccc}
\toprule
\textbf{S1} & \textbf{S2} & \textbf{S3} & \textbf{MT}
& \textbf{IoU (\%)} $\uparrow$
& \textbf{Mean CD} $\downarrow$
& \textbf{Med CD} $\downarrow$
& \textbf{Exec. (\%)} $\uparrow$ \\
\midrule
\multicolumn{8}{c}{\textit{Base model: Qwen3.5-9B}} \\
\cmark & \xmark & \xmark & \xmark
& 57.42 & 3.5819 & 0.1939 & 96.39 \\
\cmark & \cmark & \xmark & \xmark
& 54.46 & 3.4506 & 0.3520 & 93.65 \\
\cmark & \cmark & \xmark & \cmark
& 60.40 & 2.8102 & 0.1590 & 93.37 \\
\cmark & \xmark & \cmark & \cmark
& 64.26 & 2.6319 & 0.1448 & 94.28 \\
\cmark & \cmark & \cmark & \xmark
& 89.91 & 0.7492 & 0.1193 & 98.51 \\
\cmark & \cmark & \cmark & \cmark
& \textbf{91.61} & \textbf{0.5387}
& \textbf{0.1038} & \textbf{99.33} \\
\midrule
\multicolumn{8}{c}{\textit{Base model: Qwen3-VL-8B-Instruct}} \\
\cmark & \xmark & \xmark & \xmark
& 37.37 & 5.1648 & 3.9749 & 85.48 \\
\cmark & \cmark & \xmark & \cmark
& 28.30 & 5.8017 & 5.4882 & 79.87 \\
\cmark & \cmark & \cmark & \xmark
& 83.17 & 1.1057 & 0.1493 & 96.38 \\
\cmark & \cmark & \cmark & \cmark
& \textbf{85.10} & \textbf{0.8564}
& \textbf{0.1069} & \textbf{97.31} \\
\bottomrule
\end{tabular}
\end{table}

\begin{figure*}[t]
    \centering
    \includegraphics[width=\textwidth]{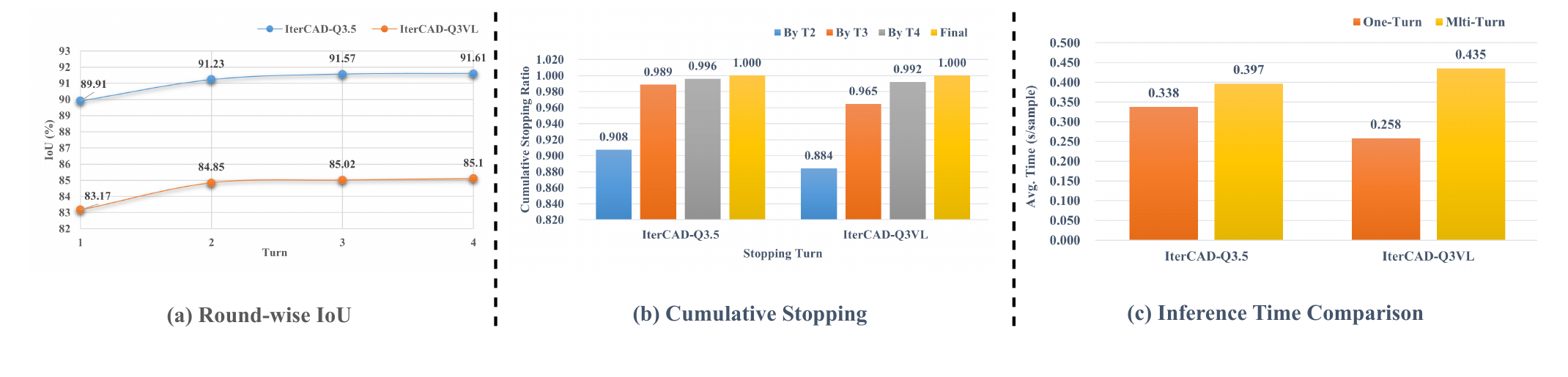}
    \caption{Iterative repair behavior on CADExpert.
    (a) Round-wise IoU.
    (b) Cumulative stopping ratios by refinement turn.
    (c) Average inference time under one-turn and multi-turn settings.}
    \Description{Three plots compare IterCAD across refinement turns.
    The first shows increasing IoU, the second shows that most samples
    stop within the early turns, and the third compares one-turn and
    multi-turn inference time for the two backbones.}
    \label{fig:multiturn_analysis}
\end{figure*}

\subsection{Ablation Studies}

Table~\ref{tab:ablation_main} reports the ablation results of \textbf{IterCAD} on CADExpert. The full configuration consistently achieves the best performance on both backbones, showing that the three-stage training recipe and multi-turn inference are all important to the final gains.

A notable observation is that the improvement cannot be explained simply by using more supervision. For example, adding Stage~II does not directly improve performance over Stage~I under single-turn inference, and can even hurt both accuracy and executability. This suggests that Stage~II introduces a more structured but also more challenging learning problem: it teaches the model to interpret intermediate CAD states and predict revise-or-stop behavior, but these abilities are not yet fully utilized when inference is restricted to a single pass.

Importantly, once multi-turn inference is enabled, the Stage~I+II model becomes clearly stronger than its single-turn counterpart. On Qwen3.5-9B, adding multi-turn inference to the Stage~I+II setting improves IoU from 54.46 to 60.40 while also reducing both mean and median Chamfer Distance. This result provides early evidence for the validity of our formulation: although Stage~II alone does not immediately improve one-shot generation, it equips the model with intermediate-state revision ability that becomes useful when the model is actually allowed to refine its prediction across turns.

Stage~III is therefore crucial. It turns locally supervised generation and revision abilities into a coherent trajectory-level repair policy. This effect is already visible in the Stage~I+III setting on Qwen3.5-9B, which outperforms Stage~I alone, indicating that trajectory-level optimization can already activate useful repair behavior. However, the full model still performs much better, showing that Stage~II and Stage~III are complementary: Stage~II provides explicit supervision on intermediate repair, while Stage~III teaches the model how to coordinate generation, revision, and stopping across turns.

Finally, comparing the full model with and without multi-turn inference shows that the learned iterative policy remains useful at test time. Enabling multi-turn inference consistently improves both geometric metrics and executability on the two backbones. This confirms that \textbf{IterCAD} does not benefit only from better training, but also from refining intermediate CAD states during inference.

\subsection{Multi-turn Inference Analysis}

To better understand how \textbf{IterCAD} behaves at test time, we analyze the quality progression, stopping behavior, and inference efficiency of multi-turn inference in Figure~\ref{fig:multiturn_analysis}. Several clear patterns can be observed.

First, iterative refinement consistently improves geometric quality across turns. As shown in Figure~\ref{fig:multiturn_analysis}(a), IoU increases monotonically for both backbones as the refinement proceeds. For IterCAD-Q3.5, IoU improves from 89.91 at Turn~1 to 91.61 at Turn~4, while for IterCAD-Q3VL it increases from 83.17 to 85.10. Most of the gain is obtained in the early turns, especially from Turn~1 to Turn~2, while later turns provide smaller but still consistent improvements. This indicates that the model is able to correct major errors early and then make finer adjustments in later rounds.

Second, the stopping behavior is highly adaptive rather than uniform. Figure~\ref{fig:multiturn_analysis}(b) shows that most samples terminate within the first two or three turns. For IterCAD-Q3.5, 90.8\% of samples stop by Turn~2 and 98.9\% by Turn~3; for IterCAD-Q3VL, the corresponding ratios are 88.4\% and 96.5\%. Only a very small fraction of samples require all four turns. This result suggests that the learned stopping policy is effective: the model does not simply run for the maximum number of steps, but instead adjusts the refinement depth according to the difficulty of the current sample.

Finally, the additional cost of multi-turn inference remains moderate. As shown in Figure~\ref{fig:multiturn_analysis}(c), the average inference time increases from 0.338\,s to 0.397\,s per sample for IterCAD-Q3.5, and from 0.258\,s to 0.435\,s for IterCAD-Q3VL, when switching from one-turn to multi-turn inference. Combined with the cumulative stopping statistics, this shows that the computational overhead of iterative repair is bounded in practice, since most samples terminate early. Overall, these results indicate that IterCAD achieves a favorable trade-off between refinement quality and inference efficiency: multi-turn inference yields consistent gains, while adaptive stopping prevents unnecessary computation.

More qualitative examples of the iterative repair process, including step-by-step reasoning and CAD refinement trajectories, are provided in the supplementary material.

\section{Conclusion}

In this paper, we presented \textbf{IterCAD}, an iterative program repair framework for CAD code generation from dimension-annotated orthographic views. Instead of treating orthographic-view-to-CAD generation as a one-shot prediction problem, IterCAD reformulates it as a multi-turn revise-or-stop process over executable intermediate CAD states. To make such iterative repair learnable, we further introduced \textbf{IterCAD-RS}, a structured revise-or-stop supervision set, together with a three-stage training recipe for initial draft generation, intermediate revision learning, and multi-turn RL optimization.

Experiments on CADExpert show that IterCAD consistently improves both geometric fidelity and code executability over strong baselines. Further ablation and multi-turn inference analyses verify that these gains come from the coordinated effect of structured intermediate supervision, trajectory-level optimization, and adaptive stopping. Overall, our results suggest that orthographic-view-to-CAD generation is better modeled as an iterative repair problem than as a one-shot generation task, and highlight the value of self-refining generation for executable CAD modeling.

\begin{acks}
This work was supported by the Program for Professor of Special
Appointment (Eastern Scholar) at Shanghai Institutions of Higher
Learning (Grant No.~GZ2022001).
\end{acks}

\bibliographystyle{ACM-Reference-Format}
\balance
\bibliography{main}


\end{document}